\documentclass[final,1p,times]{elsarticle}

\usepackage{amssymb}
\usepackage{algorithm}
\usepackage{amsmath}
\usepackage[noend]{algpseudocode}
\usepackage{subcaption}
\usepackage{multirow}
\usepackage{dsfont}
\usepackage{adjustbox,lipsum}
\usepackage{booktabs}
\journal{arxiv}

\begin{document}

\begin{frontmatter}



\title{End-to-end deep learning-based framework for path planning and collision checking: bin picking application}


\author[inst1]{Mehran Ghafarian Tamizi\corref{contrib}}
\ead{mehranght@uvic.ca}
\affiliation[inst1]{organization={Department of Electrical and Computer engineering, University of Victoria},
            addressline={3800 Finnerty Rd.}, 
            city={Victoria},
            postcode={V8P5C2}, 
            state={BC},
            country={Canada}}

\author[inst2]{Homayoun Honari\corref{contrib}}
\ead{hmnhonari@uvic.ca}
\cortext[contrib]{Authors contributed equally}
\author[inst3]{Aleksey Nozdryn-Plotnicki}
\ead{aleksey@apera.ai}
\author[inst1,inst2]{Homayoun Najjaran}
\ead{najjaran@uvic.ca}
\affiliation[inst2]{organization={Department of Mechanical engineering, University of Victoria},
            addressline={3800 Finnerty Rd.}, 
            city={Victoria},
            postcode={V8P5C2}, 
            state={BC},
            country={Canada}}
\affiliation[inst3]{organization={Apera AI},
            addressline={Ste. 501- 134 Abbott St.}, 
            city={Vancouver},
            postcode={V6B2K4}, 
            state={BC},
            country={Canada}}
\begin{abstract}
Real-time and efficient path planning is critical for all robotic systems. In particular, it is of greater importance for industrial robots since the overall planning and execution time directly impact the cycle time and automation economics in production lines. While the problem may not be complex in static environments, classical approaches are inefficient in high-dimensional environments in terms of planning time and optimality. Collision checking poses another challenge in obtaining a real-time solution for path planning in complex environments. To address these issues, we propose an end-to-end learning-based framework viz., Path Planning and Collision checking Network (PPCNet). The PPCNet generates the path by computing waypoints sequentially using two networks: the first network generates a waypoint, and the second one determines whether the waypoint is on a collision-free segment of the path. The end-to-end training process is based on imitation learning that uses data aggregation from the experience of an expert planner to train the two networks, simultaneously. We utilize two approaches for training a network that efficiently approximates the exact geometrical collision checking function. Finally, the PPCNet is evaluated in two different simulation environments and a practical implementation on a robotic arm for a bin-picking application. Compared to the state-of-the-art path planning methods, our results show significant improvement in performance by greatly reducing the planning time with comparable success rates and path lengths.
\end{abstract}




\end{frontmatter}



\section{Introduction}
\label{sec:Intro}
Bin-picking, which includes object detection, motion planning, and grasping, is a crucial part of automation in industry. Bin picking has been one of the trending research topics in recent years because of the challenges in image processing and path planning~\cite{buchholz2014combining, domae2014fast}. Industrial bin-picking consists of a 2D/3D camera, which is used to collect the environmental information (object and obstacle detection), and a conveyor with the bins. The vision system is used to detect object positions and obstacle shapes, while the motion planning module calculates the path to the grasp position based on this information~\cite{buchholz2016bin}. An efficient and real-time path for the manipulator can greatly improve the efficiency of mass production and assembly lines.

In a typical cycle of bin-picking operations, a combination of machine vision, inverse kinematics, and other algorithms are used to find the best grasping configuration for the robot. Once a configuration is found, path planning algorithms generate a path from the robot's fixed home state to the grasp, and then to a place state. Finally, the robot physically moves to complete the task. The serial and critical-path nature of path planning and movement means that planning time and execution time are essential factors that impact the overall cycle time and automation economics. It is crucial to ensure that robot paths are collision-free, considering both obstacles in the environment and self-collision.

Classical planners can be used for bin-picking, but they often fail to take advantage of the repetitive nature of the task. In a static environment, where a robot cell operates for extended periods, path planning is limited to fixed initial configurations and a fixed subspace of possible end-effector poses within a bin. This paper proposes a more efficient path planner in terms of the planning time for bin-picking tasks. We introduce the Path Planning and Collision checking Network (PPCNet), a deep neural network-based tool specifically designed for repetitive tasks like bin-picking. The PPCNet generates waypoints along the path and estimates collisions faster, making it a reliable solution for autonomous systems such as robotic arms. By using PPCNet, we can generate a safe and efficient path in a relatively short amount of time. Our proposed method outperforms conventional approaches in terms of planning time and stays competitive in path quality and success rate, making it a viable solution for industrial bin-picking applications.

The following are the major features of this research:

\begin{itemize}
    \item The proposed PPCNet end-to-end framework is an easy-to-implement and efficient approach in the context of path planning.
    \item The proposed post-processing dataset generation improves the quality of PPCNet paths.
    \item Collision detection is a critical and time-intensive aspect of path planning. PPCNet offers a faster solution, reducing planning time significantly through its ability to quickly detect collisions.
    \item We employ and compare two different methods for training the collision checking network to determine the best and safest approach for collision detection.
\end{itemize}

The paper is structured as follows: In Section~\ref{litrev}, we provide a comprehensive overview of various path planning techniques, discussing their advantages, disadvantages, and limitations for this problem. Section~\ref{pd} outlines the problem definition. We then introduce our framework, PPCNet, in Section~\ref{PPCNet}, where we also discuss the training process. In Section~\ref{result}, we evaluate and compare the performance of our approach against four state-of-the-art path planners. Finally, we conclude our study in Section~\ref{con}.

\section{Related work}\label{litrev}

\subsection{Path planning}

Generating collision-free paths for robotic arms is a key area of research in robotics. Path planning methods can broadly be classified into two categories: classical and learning-based~\cite{tamizi2023review}. Classical path planning approaches encompass a wide range of techniques, including artificial potential field (APF)~\cite{long2020virtual}, bio-inspired heuristic methods~\cite{tian2004effective}, and sampling-based path planners~\cite{elbanhawi2014sampling}. In contrast, learning-based path planners primarily utilize various machine-learning techniques to plan for the robot's path. These methods have been gaining popularity in recent years due to their ability to handle complex and dynamic environments.

Sampling-based methods are widely used in path planning. In this regard, they can be divided into two categories: Single-query algorithms and Multi-query algorithms. While single-query approaches aim to find a path between a single pair of initial and goal configurations, the multi-query ones try to be efficient when there are multiple pairs~\cite{lynch2017modern}. The Rapidly-exploring Random Tree (RRT)~\cite{lavalle1998rapidly} and Probabilistic Road Map (PRM)~\cite{latombe1998probabilistic} are two of the most well-known algorithms for single-query and multi-query approaches, respectively. PRM is a roadmap algorithm which aims to build an exhaustive roadmap of the configuration space to be able to path plan between any two configurations given to it. While PRM has been used for path planning of robotic arms before~\cite{rodriguez2014planning}, it has several limitations. Firstly, it generates paths that are far from the optimal solution. Secondly, generating the roadmap is computationally expensive~\cite{karaman2011sampling}, making it inefficient for path planning in relatively high-dimensional environments, even for static tasks such as bin-picking~\cite{ellekilde2013motion}.

Furthermore, given an initial and goal configuration pair, RRT grows a tree of waypoints from the initial configuration in order to connect the two configurations. An important characteristic of RRT algorithm is its probabilistically completeness, meaning that it is guaranteed to find a path if there exists one~\cite{lynch2017modern}. Due to RRT's  capability in non-convex high dimensional spaces, it has been extensively employed in many studies for path planning of robotic arms~\cite{rybus2015application,wang2020collision,rybus2020point}. To increase the performance of RRT in different aspects, such as planning time and optimality, there are many variants of this method in the literature. In~\cite{kuffner2000efficient}, the Bi-directional RRT (Bi-RRT) is introduced for path planning of the 7-DOF manipulator. Moreover, in~\cite{riedlinger2022concept} this algorithm used for distributed picking application. This method simultaneously builds two RRTs; one of them attempts to find the path from the initial configuration to the final configuration, and the other one attempts to find the path in the opposite direction and tries to connect these two trees in each iteration. Although this method is faster than RRT, they both often produce sub-optimal paths in practice and the quality of the path depends on the density of samples and the shape of the configuration space.

RRT* is the optimal version of the RRT and finds the asymptotically optimal solution if one exists~\cite{karaman2011sampling}. While that is a strong feature of RRT*, it is inefficient in high-dimensional spaces and has a relatively slow convergence rate. As a result, it is not a suitable solution for the real-time path planning of a robotic arm. To overcome the problems mentioned above, biased sampling is one of the promising methods which can improve the performance of sampling-based path planners, due to the adaptive sampling of the configuration space of the robot. Batch-informed trees (BIT*)~\cite{gammell2015batch} and informed RRT*~\cite{gammell2014informed} are other significant variants of the RRT* that employ an informed search strategy and batch-processing approach to find the optimal path in a more efficient way by focusing the search on promising areas of the tree. Although this method is able to find the optimal solution, it is not suitable for real-time path planning since it requires a large amount of computational resources, which can lead to delays in the system's response time.

As mentioned above, since collision-checking is computationally expensive and the configuration space gets exponentially bigger with the increase of dimensionality, most of the classical sampling-based methods suffer from high computation and low convergence rates, making them impractical for use in a complex environment with high dimensional configuration space and real-time applications such as bin picking~\cite{iversen2017benchmarking}. Ellekilde et al.~\cite{ellekilde2013motion} present a new algorithm for planning an efficient path in a bin-picking scenario based on using lookup tables. Although this method is almost instantaneous and faster than sampling-based planners, it is memory inefficient. 

To this end, learning-based path planners have emerged to deal with the limitations of classical methods. Learning-based techniques can solve the long run-time problem of sampling-based methods by providing biased sampling nodes, allowing them to run faster and more efficiently~\cite{cheng2020learning}. For instance, Qureshi et al.~\cite{qureshi2018deeply} proposed a deep neural network-based sampler to generate nodes for sampling-based path planners. This work shows a significant improvement in the performance of sampling-based path planners in terms of planning time. 
Reinforcement learning (RL) is a subcategory of learning-based techniques for manipulator path planning in an unknown environment~\cite{aleo2010sarsa,duguleana2012obstacle}. Although RL-based approaches illustrate promising performance in path planning research, they need  exhaustive interaction with the environment to acquire experience, and this is not applicable in many practical cases. 
Moreover, imitation learning is an alternative to these kinds of cases. In imitation learning, the training dataset is provided by an expert during the execution of the task. For instance, in~\cite{rahmatizadeh2016learning}, a recurrent neural network is trained by using demonstration from the virtual environment to perform a manipulation task.

Neural planners are relatively new methods in the path planning context. These planners are based on deep neural networks, and their purpose is to solve the path planning problem as an efficient and fast alternative to sampling-based methods. Bency et al.~\cite{bency2019neural} proposed a recurrent neural network to generate an end-to-end near-optimal path iteratively in a static environment. Moreover, in~\cite{qureshi2019motion} and \cite{qureshi2020motion}, motion planning networks (MPnet) is introduced. This method has a strong performance in unseen environments as the path can
be learned directly.

\subsection{Collision Approximation}

Collision detection is a vital component of path planning in robotics and can consume a significant amount of computation time. In fact, it has been reported that it can take up to 90$\%$ of the total path planning time~\cite{elbanhawi2014sampling}. There are several methods for performing collision checking in path planning. Analytical methods like GJK (Gilbert-Johnson-Keerthi) algorithm~\cite{gilbert1988fast} use mathematical equations and geometric models to determine if two objects will collide. They are often fast and efficient but can struggle with complex shapes and interactions. Grid-based Methods like Voxel Grid~\cite{lawlor2002voxel} divide the environment into a grid of cells and check for collisions by looking at the occupied cells. This is often faster than analytical methods but can result in a loss of precision. Another simple method for collision detection is by approximating objects with overlapping spheres~\cite{hubbard1996approximating}. In this approach, the number of spheres used to represent each object is a crucial hyperparameter. Additionally, this method is more instantaneous than other traditional collision detection algorithms.  

Recently, machine learning techniques have been utilized to overcome the limitations of traditional collision detection methods in robotics. For example, Support Vector Machines (SVM)~\cite{pan2015efficient} have been used to calculate precise collision boundaries in configuration space. Gaussian Mixture Models (GMM) were applied in another study~\cite{huh2016learning}, resulting in the path being generated five times faster than Bi-directional RRT. K-Nearest Neighbors (KNN) is another machine learning technique used in modeling configuration space for collision detection~\cite{pan2016fast}. Fastron, a machine learning model, was proposed by Das et al.~\cite{das2020learning} to model configuration space as a proxy collision detector, decomposing it into subspaces and training a collision checkers for each region. The majority of previous studies in this field rely on decomposing the configuration space, which requires significant engineering and simplifications.

\section{Problem definition}
\label{pd}

In this section, we will define the path-planning problem for the bin-picking application. Besides, the notation that we use in this paper will be introduced.

The first important concept in path planning is the configuration space ($C_{space}$). $C_{space}$ includes all positions and configurations of the robot. We can define $q$ as the specific configuration for an n-joints robotic arm:

\begin{equation}\label{config}
    q = \left[
\begin{array}{ccc}
\theta_{1} & ... & \theta_{n}
\end{array} 
\right]^\text{T}
\end{equation}
Where $\theta_{i}$ is the angular position of the i-th joint of the arm. $C_{free}$ can be defined as the subspace of $C_{space}$ in which the robot is collision-free. Let $\tau$ be a trajectory and the first and last element of it be $q_{initial}$ and $q_{target}$, respectively. The path planning problem is to find all of the elements between $q_{initial}$ and $q_{target}$ in a way that all of the segments in the trajectory lie entirely in the $C_{free}$ space. In the context of path planning, a segment between two configurations is the space of the configurations inside the linear combination of the two configurations. In the other words:

\begin{equation}
\Big\{\alpha q_i+(1-\alpha)q_{i+1}~|~\alpha\in[0,1]\Big\}\subseteq C_{free},  \forall i \in \{1,...,N-1\}
\end{equation}
Where N is the number of waypoints. 

\begin{figure}[htbp]
\centerline{\includegraphics[width=0.8\textwidth]{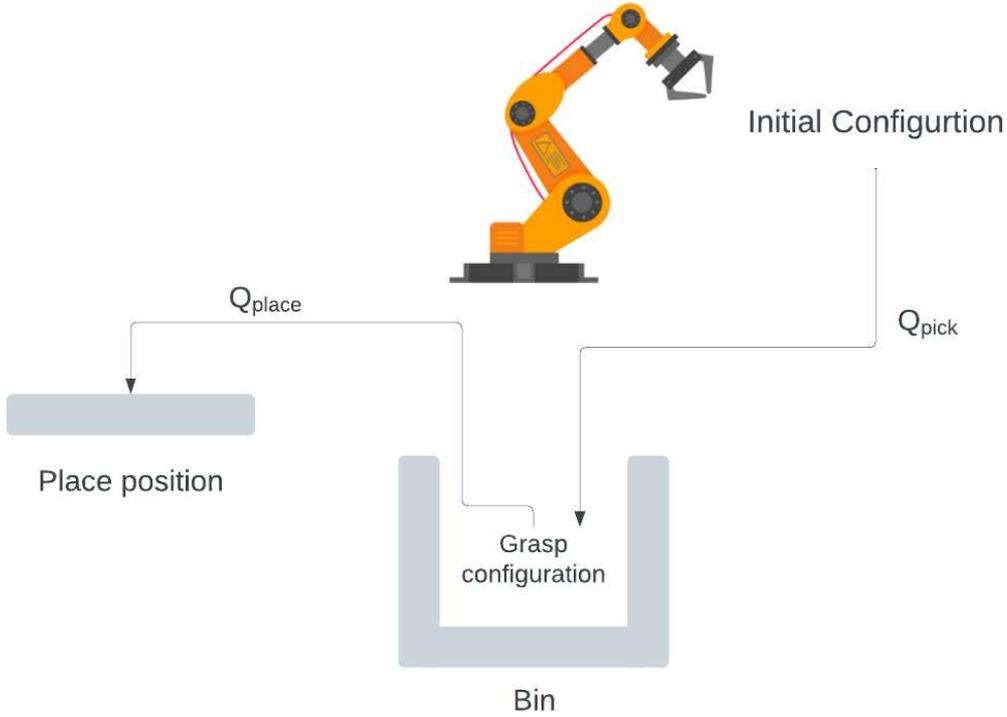}}
\caption{Path planning for pick and place operation.}
\label{overview}
\end{figure}

Bin-picking includes two different path-planning problems. The first one is finding the path between the initial configuration of the robot and the grasp configuration which is located inside the bin ($Q_{pick}$); the second one is finding the path between the grasp position and the place configuration ($Q_{place}$). Figure~\ref{overview} shows the pick-and-place operation overview. Now the problem is to design a planner that can satisfy the following objectives:
\begin{itemize}
    \item generating a safe and obstacle-free path.
    \item answering the queries in real time.
    \item generating a high-quality path with a high probability of success.
\end{itemize}

\section{Path planning and Collision checking framework}\label{PPCNet}

\begin{figure}[t]
\centering
\centerline{\includegraphics[width = 0.35\textwidth]{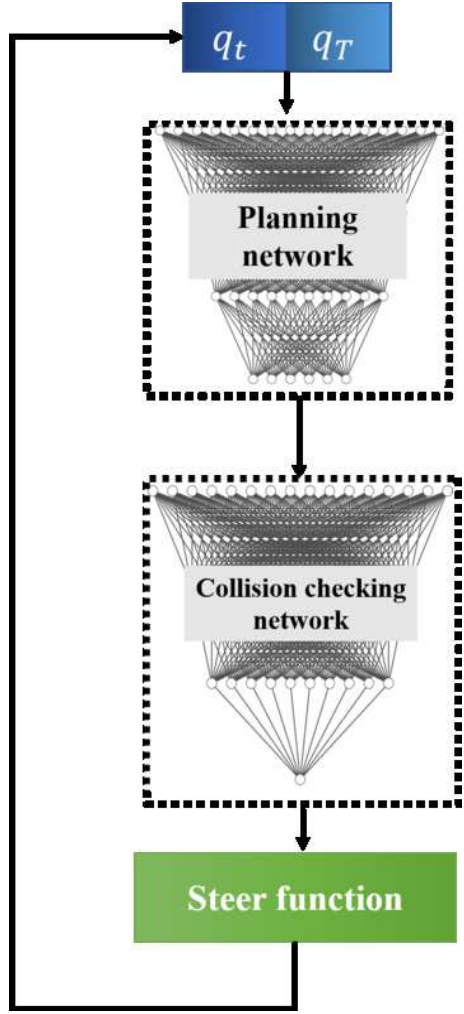}}
\caption{Path Planning and collision checking network (PPCNet).}
\label{net}
\end{figure}

In this section, we will delve into the details of the proposed framework for path planning in bin-picking tasks. As illustrated in Figure~\ref{net}, the proposed method is comprised of two consecutive networks, namely the planning network and the collision checking network. 

The first network is a modified version of MPNet~\cite{qureshi2020motion}. The original version of the MPNet consists of two networks. The first one is the planning network which generates the waypoints in a sequential manner by imitating the behavior of an expert planner. The second one, the encoder network, encodes the environment to a fixed-length vector in order to give environment information to the planning network for better planning.

The primary obstacles of concern in this paper, are the bin, other permanent structures, and the robot itself. These are considered to be fully known and encoded within the planning environment such that unlike the MPNet, our proposed planner does not use a depth map or other sensory readings of the environment to encode information for the planning network. This allows for faster planning times, which is a key focus of our research. In a full bin-picking solution, the process that determines the grasp configuration will select grasps that are at low risk of collision during grasping and robot path.

In our work, the planning network takes the current and goal configurations and generates the next joint space configuration to move into. Each configuration is represented by an n-dimensional column vector. Therefore, the path planning network's input is a 2n-dimensional vector which is obtained by the concatenation of the current and goal configuration. In each iteration of decision-making, after getting the next configuration from the path planning network, the feasibility of the segment is checked using the second network. The second network, namely the collision checking network, gets the next waypoint and estimates how likely it is that segment between the two configurations will be inside $C_{free}$. 
By following this approach, the process of decision-making (followed by collision-checking) in the configuration space can be done in a more efficient manner.

\subsection{Training and dataset generation}\label{DAGGER}

The main purpose of the path planning network is to clone the behavior of an "expert" planner. In cases like this, where it is feasible for an expert to demonstrate the desired behavior, imitation learning is a useful approach. Moreover, while RRT-based path planners are not themselves sequential decision-makers, they can be modeled as:
\begin{equation}
    q_{t+1} = f(q_t,q_{target})
\end{equation}

The purpose of the neural planner is to learn the function $f$. Behavior cloning, which focuses on using supervised learning to learn the expert's behavior, is the most basic type of imitation learning. The process of behavioral cloning is rather straightforward. The expert's demonstrations are broken down into state-action pairs, which are then used as independent and identically distributed (i.i.d.) training data for supervised learning. An important concern when using supervised learning for behavioral cloning is the risk of encountering out-of-distribution data. During the inference process, the neural planner may make decisions that were not previously observed in the dataset. Consequently, even minor inaccuracies in the replicated behavior can accumulate rapidly, resulting in a substantial failure rate in identifying viable paths. To mitigate this issue, we propose the data aggregation algorithm (DAGGER) in our study~\cite{ross2011reduction}.

The training process of the neural planner and the collision checker is explained in Algorithm \ref{dagger}. At the beginning of the process, some initial demonstrations from the expert planner is required. To do so, $RandomGoalConf(T)$ function samples T goal configurations in the collision-free space of the bin. The generated goal configurations are given to the expert planner to generate the paths using $GeneratePath$ function which requires initial and goal configurations. The expert planner outputs two datasets: the neural planner dataset ($D$) and the collision checker network dataset ($C$). $D$ only consists of the final path that the expert planner generates; however, $C$ includes every segment which was checked by the expert planner using the geometrical collision checker. In the case of an RRT-based algorithm, the final path will be stored in $D$ and the whole RRT tree and all the instances of its corresponding steer function are stored in $C$. 

\begin{figure}[tbp]
\centerline{\includegraphics[width=0.9\textwidth]{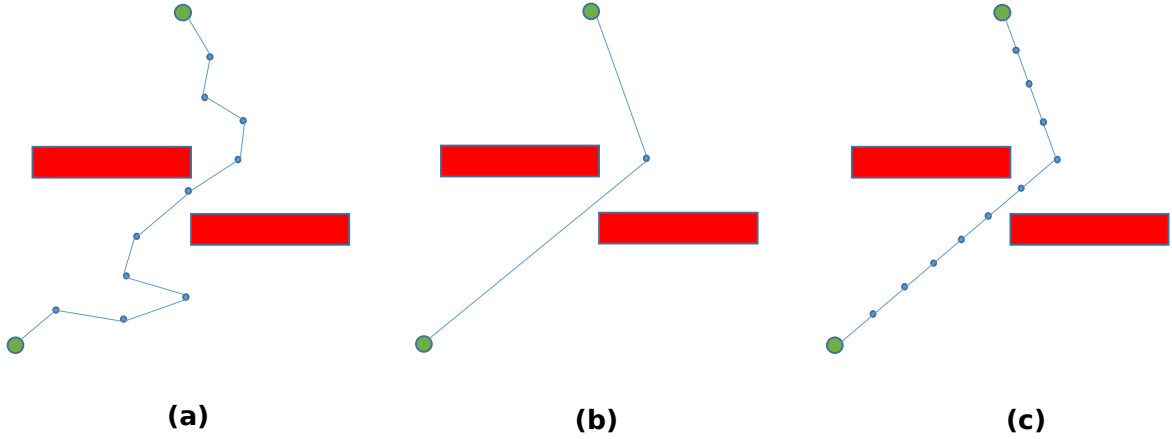}}
\caption{Post processing procedure: a) The generated path by the planner, b) Path after binary state contraction, c) Path after resampling.}
\label{resample}
\end{figure}

To train the neural planner, the demonstration dataset generated by the expert planner should be of high quality. To generate a high-quality dataset, post-processing is applied to the paths generated by the expert planner. As shown in Figure~\ref{training_fig}, the expert planner generates a collision-free path for a random query. Followed by that, the Binary State Contraction (BSC) is utilized to remove redundant and unnecessary waypoints, resulting in a shorter overall path. The sparsity of waypoints is desirable as it simplifies the subsequent processing, such as collision checking, communication with the robot controller, and trajectory planning. While the output of BSC is similar to the lazy state contraction presented in~\cite{qureshi2019motion}, BSC is computationally more efficient and reduces the run time. After BSC, the resampling function is used to ensure the smoothness and efficiency of the planned path. This involves discretizing the waypoints at regular intervals, which are guaranteed to be collision-free as they already exist on the collision-free path. Importantly, these additional waypoints do not increase the overall length of the path but do serve to reduce the mean and variance of the target step magnitudes, $||q_{t+1} - q_t||$. While larger steps require fewer neural network forward passes and collision checks during the inference phase, smaller steps offer several advantages. Firstly, they minimize the deviation from the planned path by reducing the magnitude of individual errors $||\hat{q}_{t+1} - q_{t+1}||$. Secondly, they are more likely to be collision-free as they occupy less physical space. Finally, they offer a more normalized training target for the neural network, where the problem is reduced to predicting a direction in the limit. However, BSC is a crucial step before resampling, as it relies on dividing large steps into regular steps, and a dense path of short steps cannot be divided. Figure~\ref{resample} illustrates this procedure in a 2D environment, highlighting how this method enhances the smoothness and quality of the path. The $PostProcess$ procedure in Algorithm \ref{dagger} presents the use of BSC and resampling on the generated paths.

\subsubsection{Path planner training formulation}

Once the dataset has been generated, it is time to train the network. The goal of the network is to predict the next configuration, $\hat{q}_{t+1}$, which should be as close as possible to the actual next configuration, $q_{t+1}$, in the training dataset. The network uses the dataset to learn the underlying patterns and relationships between the input and output data.

To achieve the goal of accurate prediction, the network must minimize the difference, or loss, between the predicted and actual configurations. This is done by comparing the predicted output $\hat{q}_{t+1}$ to the actual output $q_{t+1}$ and adjusting the network's parameters to minimize the difference. This process is known as backpropagation and it is typically done using optimization algorithms such as stochastic gradient descent. 

\begin{equation}
L_{planner} = \frac{1}{N_t} \sum_{j=1}^{N_d}\sum_{i=1}^{T} || \hat{q}_{j,i} - q_{j,i}||^2
\end{equation}
In this equation, $N_t$ is the number of trajectories and $N_d$ is the number of all individual waypoints in the training dataset. 

\subsubsection{Collision checker training formulation}

To ensure the safety of the robot, we experimented with two different approaches for training the collision checker and evaluated their performance. As mentioned earlier, the collision checker network takes the current and next configurations as input and estimates the probability of the segment being in-collision. Therefore, the training set for this network should include segments and corresponding labels indicating whether they are in collision or not. The optimization function utilized to train the collision checker for both of the approaches is called binary cross-entropy loss:
\begin{equation}\label{cross}
    L(\hat{p},p_{true})=\hat{p}\log{p_{true}}+(1-\hat{p})\log{(1-p_{true})}
\end{equation}
where $\hat{p}$ is the probability estimation made by the neural network and $p_{true}$ is the true probability. The reason for using binary cross entropy is due to the fact that, in essence, the collision checker is a binary classification network. Therefore, the standard approach in the literature is to use cross entropy due to it giving a better probability distribution over the training data.

\begin{figure*}[t]
\centerline{\includegraphics[width=\textwidth]{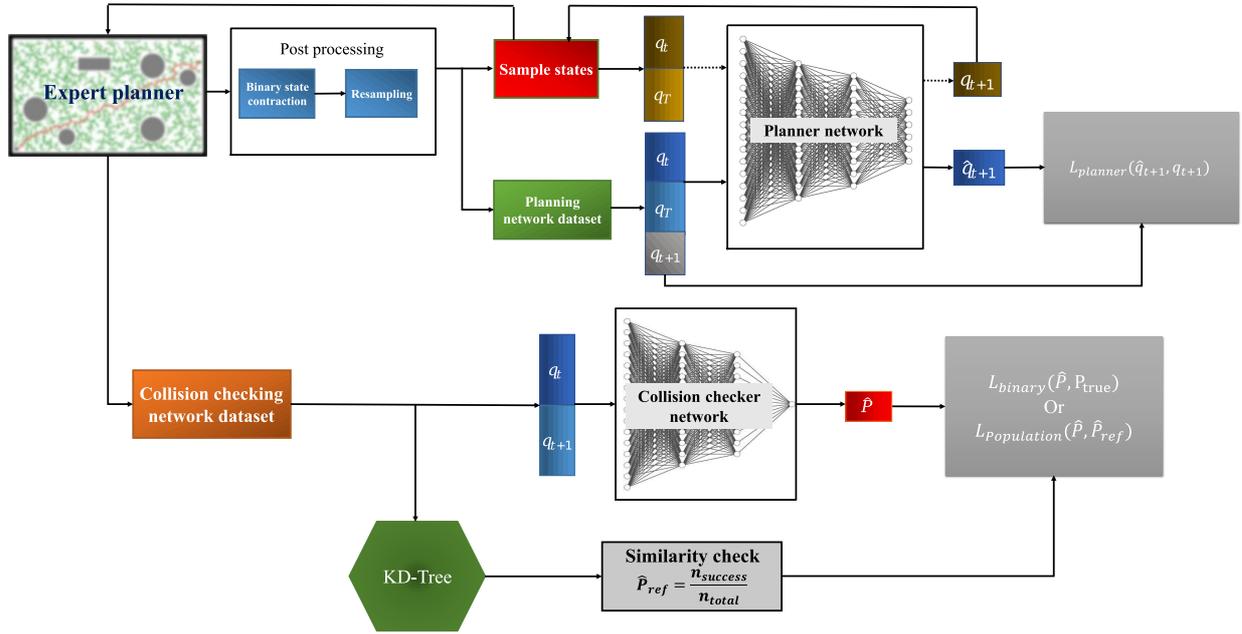}}
\caption{End-end-training process of PPCNet: \textbf{Top row:} The imitation learning and data aggregation processes for training the planner network. \textbf{Bottom row:} Population-based probability estimation and collision checker network training processes.}
\label{training_fig}
\end{figure*}

\textbf{Binary labels}: The most basic approach to tackle the optimization of the network is to use the true labels given by the analytical collision checker. In that case, $p_{true}$ in Eq. \ref{cross} becomes either 0 (collision-free segment) or 1 (in-collision segment). In Figure \ref{training_fig} this approach has been named as $L_{binary}$. The binary cross-entropy loss function measures the dissimilarity between the network's predicted probability distribution and the true probability distribution in the binary classification problem. 

\textbf{Population-based labels}: The sparsity of the binary labels hinders the performance of the collision checker since it skews the output of the network towards the more common label. The motivation for this method is to make the binary labels continuous. In addition to the mentioned reasoning, making the labels continuous can help the network be better optimized in corner cases. For example, a collision-free segment inside the bin is mostly surrounded by in-collision segments. Hence, the collision-free segment will not be of much help in optimizing the network compared to the massive existence of the in-collision segments. Making the labels continuous will amplify the effect of the collision-free segment. To make the binary labels continuous, we give a regional estimation of the probability of a specific segment being collision-free by retrieving the segments similar to it whose centers lie within the hypersphere (with a specific radius) surrounding the center of the segment. Finally, the population-based probability (denoted as $\hat{p}_{ref}$ to indicate it being a reference probability) estimates the probability of a segment being collision-free by dividing the number of collision-free segments by the total number of segments:

\begin{equation}
    \hat{p}_{ref}=\frac{\sum_{k=1}^K\mathds{1}[p_{true}^{(k)}==0]}{K}
\end{equation}

where $\mathds{1}$ is the identity-indication function. The advantage of the proposed approach is that the probability labels generated by this method can provide a well-behaving continuous function where near the obstacles, the probability gets near zero and the further it gets from them, it approaches one. In practice, we utilize KD-Tree method (built with the centers of the segments) to find similar segments. Moreover, the more data this approach gets, the better the estimation will be. Therefore, our proposed algorithm stores every segment that was checked by the analytical collision checker during the path-generation process with the expert planner. In the case of an RRT-based expert planner, this corresponds to storing the whole tree and all of the in-collision segments generated by it. While this approach may have higher computational cost compared to the binary labels approach, our results indicate that it can achieve better prediction performance. For this approach, $\hat{p}_{ref}$ is used instead of $p_{true}$ in 
Eq. \ref{cross} and the corresponding loss function is called $L_{population}$.

\begin{algorithm}[t]
\caption{Training process of PPCNet}\label{dagger}
\centering
\begin{algorithmic}
\State \textbf{Require} expert planner $\pi^*$, number of initial demonstrations $T$, number of rollouts in each iteration  $T'$,number of state samples $S$, required policy success rate $\zeta$
\State Initialize Planner and Collision Dataset: $D,C\gets \emptyset$
\State \text{Generate initial demonstrations:}
\State \quad $GoalConf\gets RandomGoalConf(T)$
\State \quad$D,C \gets GeneratePath(\pi^*,HomeConf,GoalConf)$
\State Apply post-processing: $PostProcess(D)$
\State $SuccessRate \gets 0$
\For{$i=1,...$}
    \State $\pi_i \gets TrainPolicy(D)$
    \Comment{Start of DAGGER}
    \State $GoalConf \gets RandomGoalConf(T')$
    \State $D_{\pi_i} \gets GeneratePath(\pi_i,HomeConf,GoalConf)$
    \State $S_i \gets SampleStates(D_{\pi_i},S)$
    
    \State $D_i,C_i \gets GeneratePath(\pi^*,S_i,GoalConf)$
    \State $D_i \gets PostProcess(D_i)$
    \State $D \gets D \cup D_i$.
    \Comment{End of DAGGER}
    \State $C \gets C \cup C_i$.
    \State $\eta_i \gets TrainCollisionChecker(C)$
    \State $SuccessRate\gets TestPolicy(\pi_i,\eta_i)$
    \If{$SuccessRate>\zeta$}
    \State \Return $\pi_i,\eta_i$
    \EndIf

\EndFor
\end{algorithmic}
\end{algorithm}

\subsubsection{End-to-end training process}

In this section, an overview of the end-to-end process of training the integrated model which consists of both the planner network and the collision checker network will be discussed. The training procedure is outlined in Algorithm \ref{DAGGER}.

Initially, a set of trajectories are sampled using the expert planner as the initial demonstration, and the planner and collision dataset are filled. In each iteration, the path planner's policy is trained using the demonstrations from an expert planner. Consequently, DAGGER process (as explained in Section \ref{DAGGER}) is executed which results in the expansion of both of the datasets. Finally, the collision-checking dataset is trained and the model's performance is evaluated in the $TestPolicy$ function. The function uses some random goal configurations and attempts to path plan between them (following the process in Algorithm \ref{pickplanner}). If the success rate of the model was satisfying, the path planner and the collision checker network weights are returned.

\begin{algorithm}[t]
\caption{Path generation process using PPCNet}\label{pickplanner}
\begin{algorithmic}
\State \textbf{Require} Trained neural planner $\pi$ and collision checker $\eta$, and initial and goal configurations $q_{start},q_{target}$

\State $path\gets \{q_{start}\}$
\State $q_{current} \gets q_{start}$
\For{$i=1,...,s_{max}$}
\If{$Steer(q_{current},q_{target},\eta)==q_{target}$}
    \State $path\gets path \cup q_{target}$
    \If{$IsFeasible(path)$}
    \State \Return $path$
    \Else
    \Comment{Apply Patching}
    \State $segments\gets InCollision(path)$
    \For{$(q_s,q_e) ~in ~segments$}
        \State $path_{alt}\gets \pi^*(q_s,q_e)$
        \If{$Failed(path_{alt})$}
            \State \Return failure
        \EndIf
        \State $path\gets Patch(path,path_{alt})$
    \EndFor
    \State \Return $path$
    \EndIf
\EndIf
\State $q_{next} \gets \pi(q_{current},q_{target})$
\If{$Steer(q_{current},q_{next},\eta)$}
    \State $q_{next} \gets Steer(q_{current},q_{next},\eta)$
    \State $path\gets path\cup q_{next}$
    \State $q_{current}\gets q_{next}$
\EndIf
\EndFor
\end{algorithmic}
\end{algorithm}

\begin{algorithm}[b]
\caption{Steer Function $(q_{current},q_{next})$}\label{steer}
\begin{algorithmic}
\State \textbf{Require} collision checking network $\eta$, safety threshold $\Bar{\eta}$
\State $segments\gets Discretize(q_{current},q_{next})$
\State $q_{final} \gets q_{current}$
\For{$(q_i,q_e) ~in ~segments$}
    \If{$\eta(q_i,q_e)>\Bar{\eta}$}
        \State $q_{final}\gets q_e$
    \Else
        \If{$q_{final}==q_{current}$}
            \State \Return failure
        \Else
            \State \Return success, $q_{final}$
        \EndIf
    \EndIf
\EndFor

\end{algorithmic}
\end{algorithm}
\subsubsection{Path planning process}

In this section, the end-to-end path planning process of the PPCNet is explained. As illustrated in Figure \ref{net}, the two networks, the planning network and collision checking network, do the decision-making sequentially. In the context of the bin-picking application, given the start, pick, and place configurations, the proposed model picks the object from the bin and places it in the place configuration.
As outlined in Algorithm \ref{pickplanner}, in each iteration of the decision-making process the path planning network outputs the next configuration that the robotic arm must transition into. The current configuration and the next configuration are then checked with the collision-checking network. To do so, the $Steer$ function is used. 
As explained in Algorithm \ref{steer}, the path between the two waypoints is discretized into equal segments with lengths close to the resolution step size of the expert planner. This is done in order to have the segments be close to the collision checker training dataset's distribution. Furthermore, in each segment, the collision-free probability of the segment is checked. In order to decide whether a segment is collision-free or not, a safety threshold is needed. If the probability is more the threshold, the algorithm continues to the next segment for collision checking. If the probability is less than the threshold, then the segment is deemed to be in-collision. In that case, if the in-collision segment is the first segment of the path, then the planner has failed to output a good waypoint. Otherwise, if the in-collision segment is not the first segment, the planning network was successful and the tail of the last collision-free segment will be the output of the $steer$ function.
Moving on, if the $steer$ function outputs success, then the next waypoint toward the goal configuration is added to the path. In the case that the steer function outputs failure, the planner network attempts to generate a new waypoint. For that purpose, stochasticity is needed inside the network. Therefore, dropout layers are used during inference to have the network be able to recover from failure cases~\cite{qureshi2019motion}.
Furthermore, in each iteration, if it is possible to directly connect the current configuration to the target configuration, then the target configuration is added to the path and it will be the final path generated by the PPCNet.
At this point, the feasibility of the path is checked with the geometrical collision checker using $IsFeasible$ function. If the path was collision-free, it will be returned; otherwise, the algorithm will attempt to patch the in-collision segments using the expert planner. To this end, the $InCollision$ function outputs the segments in the path that are in-collision. It is important to note that if some consecutive segments were in-collision, all of the segments will be counted as one bigger in-collision segment and the head and tail of the bigger segment will be considered. This is done in order to have lesser expert planner calls and be more computationally efficient. Furthermore, for each segment, the expert planner attempts to generate an alternative path between the two configurations. If for all of the segments, the patching was done with success, then the proposed path planner would be successful and the path would be returned.

\begin{figure}[t]
\centerline{\includegraphics[width=0.7\textwidth]{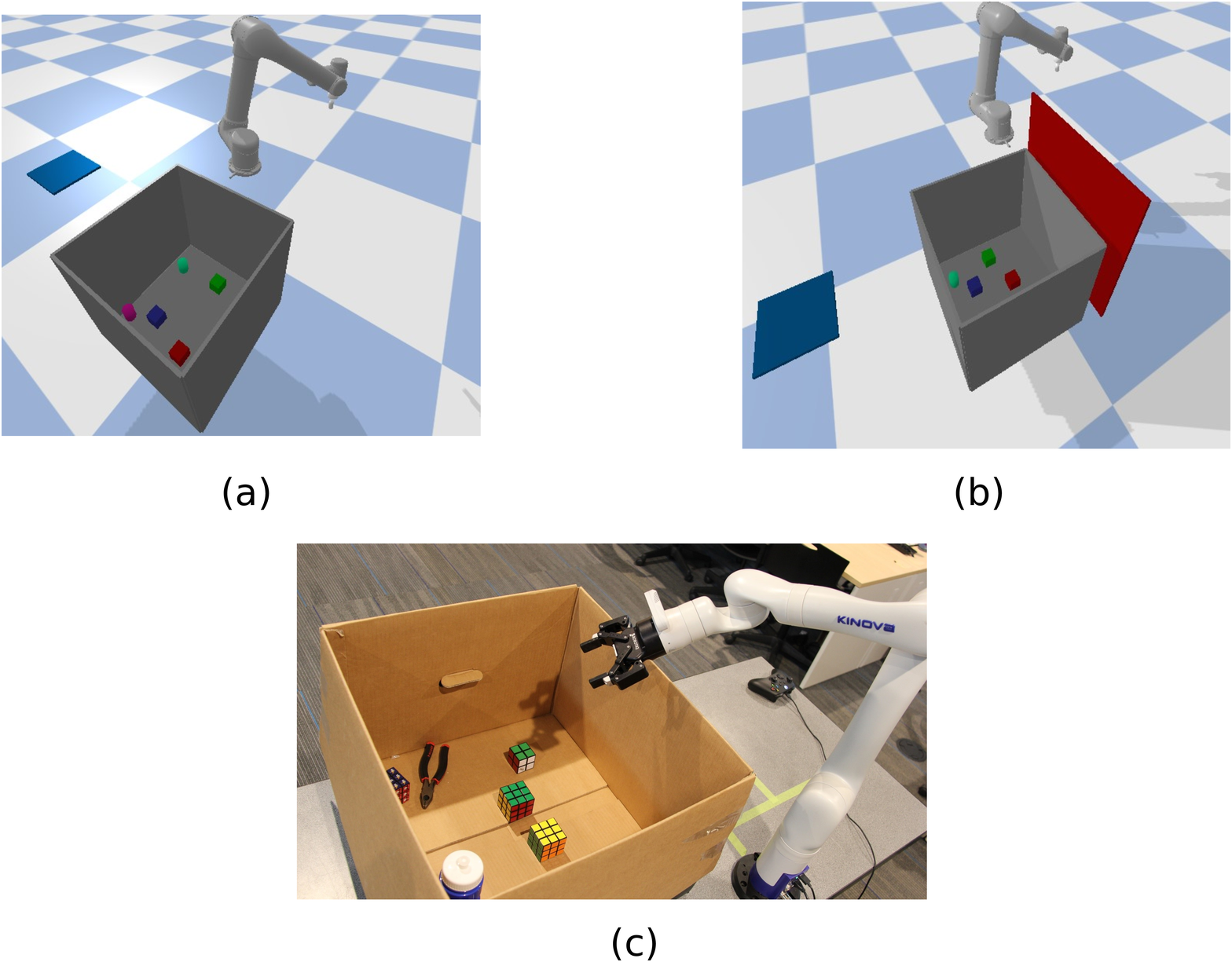}}
\caption{Experimental environments: a) UR5 scene, b) UR5 scene with a wall as an obstacle, c) Real-world implementation on Kinova Gen3 .}
\label{env}
\end{figure}

\begin{table}[t]
\centering
\caption{Hyperparameters selection for planners.}
\label{hyper}
\begin{adjustbox}{ width=0.6\textwidth}
\begin{tabular}{ccc}
\hline
Methods       & Hyperparameters                                                                                               & Values                                                                                       \\ \hline
              & Max iterations                                                                                                & 1000                                                                                         \\
Bi-RRT        & Max planning time                                                                                             & 5 s                                                                                          \\
              & Resolution                                                                                                    & 0.1 rad                                                                                      \\ \hline
              & Number of batches                                                                                             & 500                                                                                          \\
              & Resolution                                                                                                    & 0.05                                                                                         \\
Informed RRT* & $\gamma$                                                                                                      & 500                                                                                          \\
              & Goal probability                                                                                              & 0.1                                                                                     \\
              & Max iterations                                                                                                & 1000                                                                                         \\
              & Max planning time                                                                                             & 10 s                                                                                         \\ \hline
              & Max iterations                                                                                                & 1000                                                                                         \\
              & Max planning time                                                                                             & 10 s                                                                                         \\
BIT*          & Number of samples                                                                                             & 500                                                                                          \\
              & $\eta$                                                                                                        & 20                                                                                           \\
              & Goal probability                                                                                              & 0.1                                                                                          \\ \hline
              & Max iterations                                                                                                & 100                                                                                          \\
              & learning rate                                                                                                 & 0.0001                                                                                       \\
MPNet         & Expert planner                                                                                                & Bi-RRT                                                                                       \\
              & Network architecture                                                                                          & \begin{tabular}[c]{@{}c@{}}Fully connected\\ 6 layers, 300 neurons \\ per layer\end{tabular} \\ \hline
              & \begin{tabular}[c]{@{}c@{}}Post-processing \\ resample step size\end{tabular}                                 & 0.1745 rad                                                                                   \\
              & Max iterations                                                                                                & 100                                                                                          \\
              & Optimizer                                                                                                     & Adam                                                                                         \\
PPCNet        & Similarity distance                                                                                           & 0.4 rad                                                                                      \\
              & Dagger iterations                                                                                             & 30                                                                                           \\
              & Expert planner                                                                                                & Bi-RRT                                                                                       \\
              & Safety threshold ($\bar{\eta}$)                                                                               & 0.8                                                                                          \\
              & \begin{tabular}[c]{@{}c@{}}Network architecture\\ (planning and , collision \\ checking network)\end{tabular} & \begin{tabular}[c]{@{}c@{}}Fully connected\\ 6 layers, 300 neurons \\ per layer\end{tabular} \\ \hline

\end{tabular}
\end{adjustbox}
\end{table}

\section{Experimental results}\label{result}

In this section, we will discuss the performance of our proposed algorithm, which employs PyTorch~\cite{paszke2017automatic} to implement the planner network and collision checker network. The simulation environment we used is Pybullet planning~\cite{pyplanning, coumans2021}. To evaluate the effectiveness of our approach compared to the state-of-the-art path planning techniques, we compared it against Bi-RRT~\cite{kuffner2000efficient}, informed RRT*~\cite{gammell2014informed}, BIT*~\cite{gammell2015batch}, and MPNet~\cite{qureshi2020motion} on three different environments (Figure~\ref{env}). The first environment consists of a universal robot UR5 and a bin, the second one involves a UR5 robot and a wall as an obstacle, and the third one features a Kinova Gen3 robotic arm, which we implemented on both simulation and practical settings.  The system used for training and testing has a 3.700 GHz AMD Ryzen 9 processor with 96 GB RAM and NVIDIA TITAN RTX GPU. Moreover, Table~\ref{hyper} shows the hyperparameter selection for each of the planners. 

\begin{figure}[h]
\centerline{\includegraphics[width=1.2\textwidth]{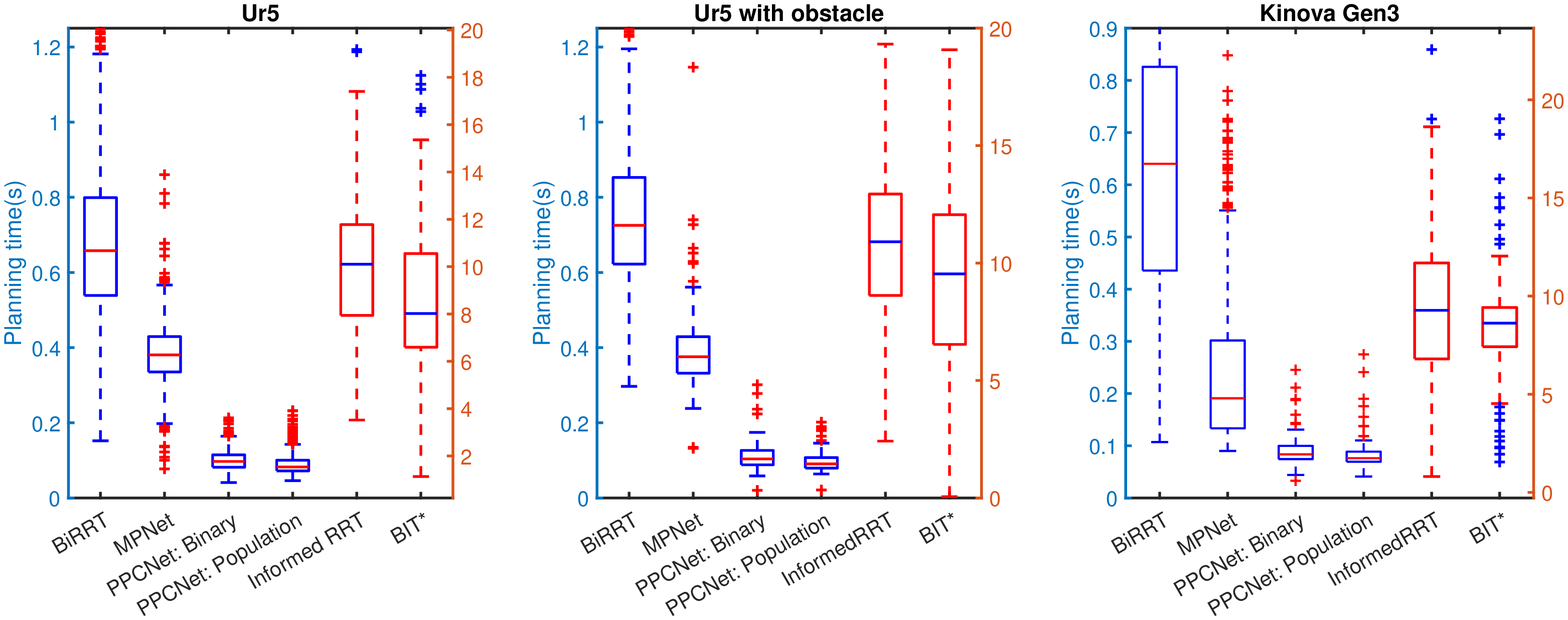}}
\caption{Planning time comparison between different methods.}
\label{time}
\end{figure}

Table~\ref{compare} presents a comparative analysis of the performance of PPCNet against other planning methods. Our experiments demonstrate that PPCNet exhibits remarkable performance in terms of planning time in all three environments. Figure~\ref{time} provides a visual illustration of the planning time comparison. The hyperparameters for the competing RRT-based algorithms were tuned such that they produce paths of comparable length to PPCNet, while ensuring that they find a path for each planning task in less than 10 seconds and 1000 iterations. However, the results show that BIT* and Informed RRT* methods take longer to plan and have lower success rates than PPCNet.

PPCNet has shown its ability to generate paths with comparable length and success rates compared to other algorithms, which is particularly important in evaluating the effectiveness of path-planning methods. Additionally, the post-processing method employed in dataset generation has allowed PPCNet to outperform its expert planner (Bi-RRT) in terms of path length. The BSC function plays a critical role in eliminating redundant and unnecessary waypoints, resulting in a reduction in path length and an improvement in path quality. The combination of the BSC function and resampling has significantly enhanced the quality of paths generated by PPCNet. Furthermore, our findings show that while the addition of the collision checking to the PPCNet has caused significant improvement compared to the MPNet in the expense of the marginal reduction in the PPCNet's success rate.

Finally, the study's results indicate that the population-based training method for the collision-checking network performed better than the binary classification approach, resulting in a more accurate collision-checking model. This is evident from the decrease in the planning time due to a reduction in the need for patching. Hence, the trade-off between the computational cost of training the population-based collision checker against the accuracy it offers can be observed.

\begin{table}[t]
\centering
\caption{Planning time and path length comparison of the proposed method and BI-RRT for 500 random pick-and-place queries.}
\label{compare}
\begin{adjustbox}{max width=\textwidth}
\begin{tabular}{@{}ccccc@{}}
\toprule
\textbf{Environments} & \textbf{Methods}                                                       & \textbf{Planning time (s)} & \textbf{Path length (rad)} & \textbf{Success rate ($\%$)} \\ \midrule
                     & Bi-RRT                                                                 & 0.701 ± 0.246              & 7.516 ± 2.147              & 100                          \\
                     & Informed RRT*                                                          & 10.186 ± 3.192             & 5.304 ± 1.983              & 79.8                         \\
                     & BIT*                                                                   & 8.621 ± 3.594              & 5.278±2.315                & 77.6                         \\
UR5                  & MPNet                                                                  & 0.384 ± 0.079              & 7.305 ± 2.298              & 94.4                         \\
                     & \begin{tabular}[c]{@{}c@{}}PPCNet\\ Collision: Binary\end{tabular}     & 0.101 ± 0.031              & 5.887 ± 1.038              & 90.2                         \\
                     & \begin{tabular}[c]{@{}c@{}}PPCNet\\ Collision: Population\end{tabular} & 0.093 ± 0.033              & 5.679 ± 1.503              & 93.4                         \\ \midrule
                     & Bi-RRT                                                                 & 0.749 ± 0.182              & 7.925 ± 3.307              & 100                          \\
                     & Informed RRT*                                                          & 10.856 ± 3.425             & 5.813 ± 2.056              & 73.6                         \\
                     & BIT*                                                                   & 9.299 ± 3.919              & 5.514 ± 2.747              & 70.4                         \\
UR5 with a wall    & MPNet                                                                  & 0.392 ± 0.106              & 7.874 ± 3.011              & 93                           \\
                     & \begin{tabular}[c]{@{}c@{}}PPCNet\\ Collision: Binary\end{tabular}     & 0.106 ± 0.027              & 6.142 ± 1.979              & 89.8                         \\
                     & \begin{tabular}[c]{@{}c@{}}PPCNet\\ Collision: Population\end{tabular} & 0.099 ± 0.034              & 6.016 ± 1.749              & 92.2                         \\ \midrule
                     & Bi-RRT                                                                 & 0.645 ± 0.281              & 6.812 ± 2.749              & 100                          \\
                     & Informed RRT*                                                          & 9.535 ± 4.122              & 5.124 ± 2.205              & 79.2                         \\
                     & BIT*                                                                   & 8.496 ± 3.082              & 5.099 ± 2.464              & 79.6                         \\
Kinova Gen3          & MPNet                                                                  & 0.227 ± 0.151              & 6.714 ± 2.812              & 90.6                         \\
                     & \begin{tabular}[c]{@{}c@{}}PPCNet\\ Collision: Binary\end{tabular}     & 0.090 ± 0.029              & 5.315 ± 1.482              & 88.2                         \\
                     & \begin{tabular}[c]{@{}c@{}}PPCNet\\ Collision: Population\end{tabular} & 0.085 ± 0.032              & 5.188 ± 1.335              & 89.8                         \\ \bottomrule
\end{tabular}
\end{adjustbox}
\end{table}

\section{Conclusions}\label{con}
In this paper, we introduced a novel path-planning framework, PPCNet, and examined its efficacy for bin-picking applications. The robot path planning for the bin-picking application is typically a computationally expensive process because the traditional algorithms need to compute the path in its entirety every time the target changes. Further, most existing algorithms lack the ability to learn from past experiences. The proposed PPCNet utilizes deep neural networks to find a real-time solution for path-planning and collision-checking problems. Specifically, we propose an end-to-end training algorithm that leverages the data aggregation algorithm to generate i.i.d. data for supervised learning. Using an expert planner in the training process, data aggregation can assist the agent in encountering unseen states. Another novel component of the proposed PPCNet algorithm includes two variations of the collision-checking network for approximating the exact geometrical collision function. We evaluate our proposed framework, PPCNet, by conducting simulation on two robotic arms namely UR5 and Kinova Gen3 in a bin-picking application. We have also examined the algorithm in a real-world scenario using Kinova Gen3 robot. Our results indicate that our approach significantly reduced planning time while producing paths with path lengths and success rates comparable to those of the state-of-the-art path planning methods.

\section*{Acknowledgment}
We would like to acknowledge the financial support of Apera AI and Mathematics of Information Technology and Complex Systems (MITACS) under IT16412 Mitacs Accelerate.

 \bibliographystyle{elsarticle-num} 
 \bibliography{cas-refs}

\begin{thebibliography}{10}
\expandafter\ifx\csname url\endcsname\relax
  \def\url#1{\texttt{#1}}\fi
\expandafter\ifx\csname urlprefix\endcsname\relax\def\urlprefix{URL }\fi
\expandafter\ifx\csname href\endcsname\relax
  \def\href#1#2{#2} \def\path#1{#1}\fi

\bibitem{buchholz2014combining}
D.~Buchholz, D.~Kubus, I.~Weidauer, A.~Scholz, F.~M. Wahl, Combining visual and
  inertial features for efficient grasping and bin-picking, in: 2014 IEEE
  international conference on robotics and automation (ICRA), IEEE, 2014, pp.
  875--882.

\bibitem{domae2014fast}
Y.~Domae, H.~Okuda, Y.~Taguchi, K.~Sumi, T.~Hirai, Fast graspability evaluation
  on single depth maps for bin picking with general grippers, in: 2014 IEEE
  International Conference on Robotics and Automation (ICRA), IEEE, 2014, pp.
  1997--2004.

\bibitem{buchholz2016bin}
D.~Buchholz, Bin-picking, Studies in Systems, Decision and Control 44 (2016)
  3--12.

\bibitem{tamizi2023review}
M.~G. Tamizi, M.~Yaghoubi, H.~Najjaran, A review of recent trend in motion
  planning of industrial robots, International Journal of Intelligent Robotics
  and Applications (2023) 1--22.

\bibitem{long2020virtual}
Z.~Long, Virtual target point-based obstacle-avoidance method for manipulator
  systems in a cluttered environment, Engineering Optimization 52~(11) (2020)
  1957--1973.

\bibitem{tian2004effective}
L.~Tian, C.~Collins, An effective robot trajectory planning method using a
  genetic algorithm, Mechatronics 14~(5) (2004) 455--470.

\bibitem{elbanhawi2014sampling}
M.~Elbanhawi, M.~Simic, Sampling-based robot motion planning: A review, Ieee
  access 2 (2014) 56--77.

\bibitem{lynch2017modern}
K.~M. Lynch, F.~C. Park, Modern robotics, Cambridge University Press, 2017.

\bibitem{lavalle1998rapidly}
S.~M. LaValle, et~al., Rapidly-exploring random trees: A new tool for path
  planning, Tech. rep. (1998).

\bibitem{latombe1998probabilistic}
L.~E. K. J.-C. Latombe, Probabilistic roadmaps for robot path planning,
  Practical motion planning in robotics: current approaches and future
  challenges, Citeseer (1998) 33--53.

\bibitem{rodriguez2014planning}
C.~Rodr{\'\i}guez, A.~Monta{\~n}o, R.~Su{\'a}rez, Planning manipulation
  movements of a dual-arm system considering obstacle removing, Robotics and
  Autonomous Systems 62~(12) (2014) 1816--1826.

\bibitem{karaman2011sampling}
S.~Karaman, E.~Frazzoli, Sampling-based algorithms for optimal motion planning,
  The international journal of robotics research 30~(7) (2011) 846--894.

\bibitem{ellekilde2013motion}
L.-P. Ellekilde, H.~G. Petersen, Motion planning efficient trajectories for
  industrial bin-picking, The International Journal of Robotics Research
  32~(9-10) (2013) 991--1004.

\bibitem{rybus2015application}
T.~Rybus, K.~Seweryn, Application of rapidly-exploring random trees (rrt)
  algorithm for trajectory planning of free-floating space manipulator, in:
  2015 10th International Workshop on Robot Motion and Control (RoMoCo), IEEE,
  2015, pp. 91--96.

\bibitem{wang2020collision}
X.~Wang, X.~Luo, B.~Han, Y.~Chen, G.~Liang, K.~Zheng, Collision-free path
  planning method for robots based on an improved rapidly-exploring random tree
  algorithm, Applied Sciences 10~(4) (2020) 1381.

\bibitem{rybus2020point}
T.~Rybus, Point-to-point motion planning of a free-floating space manipulator
  using the rapidly-exploring random trees (rrt) method, Robotica 38~(6) (2020)
  957--982.

\bibitem{kuffner2000efficient}
J.~Kuffner, S.~L. RRT-Connect, An efficient approach to single-query path
  planning ieee international conference on robotics and automation, San
  Francisco (2000) 473--479.

\bibitem{riedlinger2022concept}
M.~A. Riedlinger, M.~G. Tamizi, J.~Tikekar, M.~Redeker, Concept for a
  distributed picking application utilizing robotics and digital twins, 27th
  IEEE ETFA (2022).

\bibitem{gammell2015batch}
J.~D. Gammell, S.~S. Srinivasa, T.~D. Barfoot, Batch informed trees (bit):
  Sampling-based optimal planning via the heuristically guided search of
  implicit random geometric graphs, in: 2015 IEEE international conference on
  robotics and automation (ICRA), IEEE, 2015, pp. 3067--3074.

\bibitem{gammell2014informed}
J.~D. Gammell, S.~S. Srinivasa, T.~D. Barfoot, Informed rrt*: Optimal
  sampling-based path planning focused via direct sampling of an admissible
  ellipsoidal heuristic, in: 2014 IEEE/RSJ International Conference on
  Intelligent Robots and Systems, IEEE, 2014, pp. 2997--3004.

\bibitem{iversen2017benchmarking}
T.~F. Iversen, L.-P. Ellekilde, Benchmarking motion planning algorithms for
  bin-picking applications, Industrial Robot: An International Journal (2017).

\bibitem{cheng2020learning}
R.~Cheng, K.~Shankar, J.~W. Burdick, Learning an optimal sampling distribution
  for efficient motion planning, in: 2020 IEEE/RSJ International Conference on
  Intelligent Robots and Systems (IROS), IEEE, 2020, pp. 7485--7492.

\bibitem{qureshi2018deeply}
A.~H. Qureshi, M.~C. Yip, Deeply informed neural sampling for robot motion
  planning, in: 2018 IEEE/RSJ International Conference on Intelligent Robots
  and Systems (IROS), IEEE, 2018, pp. 6582--6588.

\bibitem{aleo2010sarsa}
I.~Aleo, P.~Arena, L.~Patan{\'e}, Sarsa-based reinforcement learning for motion
  planning in serial manipulators, in: The 2010 International Joint Conference
  on Neural Networks (IJCNN), IEEE, 2010, pp. 1--6.

\bibitem{duguleana2012obstacle}
M.~Duguleana, F.~G. Barbuceanu, A.~Teirelbar, G.~Mogan, Obstacle avoidance of
  redundant manipulators using neural networks based reinforcement learning,
  Robotics and Computer-Integrated Manufacturing 28~(2) (2012) 132--146.

\bibitem{rahmatizadeh2016learning}
R.~Rahmatizadeh, P.~Abolghasemi, A.~Behal, L.~B{\"o}l{\"o}ni, Learning real
  manipulation tasks from virtual demonstrations using lstm, arXiv preprint
  arXiv:1603.03833 (2016).

\bibitem{bency2019neural}
M.~J. Bency, A.~H. Qureshi, M.~C. Yip, Neural path planning: Fixed time,
  near-optimal path generation via oracle imitation, in: 2019 IEEE/RSJ
  International Conference on Intelligent Robots and Systems (IROS), IEEE,
  2019, pp. 3965--3972.

\bibitem{qureshi2019motion}
A.~H. Qureshi, A.~Simeonov, M.~J. Bency, M.~C. Yip, Motion planning networks,
  in: 2019 International Conference on Robotics and Automation (ICRA), IEEE,
  2019, pp. 2118--2124.

\bibitem{qureshi2020motion}
A.~H. Qureshi, Y.~Miao, A.~Simeonov, M.~C. Yip, Motion planning networks:
  Bridging the gap between learning-based and classical motion planners, IEEE
  Transactions on Robotics 37~(1) (2020) 48--66.

\bibitem{gilbert1988fast}
E.~G. Gilbert, D.~W. Johnson, S.~S. Keerthi, A fast procedure for computing the
  distance between complex objects in three-dimensional space, IEEE Journal on
  Robotics and Automation 4~(2) (1988) 193--203.

\bibitem{lawlor2002voxel}
O.~S. Lawlor, L.~V. Kal{\'e}e, A voxel-based parallel collision detection
  algorithm, in: Proceedings of the 16th international conference on
  Supercomputing, 2002, pp. 285--293.

\bibitem{hubbard1996approximating}
P.~M. Hubbard, Approximating polyhedra with spheres for time-critical collision
  detection, ACM Transactions on Graphics (TOG) 15~(3) (1996) 179--210.

\bibitem{pan2015efficient}
J.~Pan, D.~Manocha, Efficient configuration space construction and optimization
  for motion planning, Engineering 1~(1) (2015) 046--057.

\bibitem{huh2016learning}
J.~Huh, D.~D. Lee, Learning high-dimensional mixture models for fast collision
  detection in rapidly-exploring random trees, in: 2016 IEEE International
  Conference on Robotics and Automation (ICRA), IEEE, 2016, pp. 63--69.

\bibitem{pan2016fast}
J.~Pan, D.~Manocha, Fast probabilistic collision checking for sampling-based
  motion planning using locality-sensitive hashing, The International Journal
  of Robotics Research 35~(12) (2016) 1477--1496.

\bibitem{das2020learning}
N.~Das, M.~Yip, Learning-based proxy collision detection for robot motion
  planning applications, IEEE Transactions on Robotics 36~(4) (2020)
  1096--1114.

\bibitem{ross2011reduction}
S.~Ross, G.~Gordon, D.~Bagnell, A reduction of imitation learning and
  structured prediction to no-regret online learning, in: Proceedings of the
  fourteenth international conference on artificial intelligence and
  statistics, JMLR Workshop and Conference Proceedings, 2011, pp. 627--635.

\bibitem{paszke2017automatic}
A.~Paszke, S.~Gross, S.~Chintala, G.~Chanan, E.~Yang, Z.~DeVito, Z.~Lin,
  A.~Desmaison, L.~Antiga, A.~Lerer, Automatic differentiation in pytorch
  (2017).

\bibitem{pyplanning}
C.~R. Garrett, \href{https://pypi.org/project/pybullet-planning/}{Pybullet
  planning} (2020).
\newline\urlprefix\url{https://pypi.org/project/pybullet-planning/}

\bibitem{coumans2021}
E.~Coumans, Y.~Bai, Pybullet, a python module for physics simulation for games,
  robotics and machine learning, \url{http://pybullet.org} (2016--2021).

\end{thebibliography}





\end{document}